\pdfoutput=1

\documentclass[11pt]{article}

\usepackage[]{emnlp2021}

\usepackage{times}
\usepackage{latexsym}

\usepackage{subcaption}
\usepackage{graphicx}

\usepackage[T1]{fontenc}

\usepackage[utf8]{inputenc}

\usepackage{microtype}

\usepackage{multirow}

\usepackage{comment}
\usepackage{macro}
\usepackage{color}

\title{T\textsuperscript{3}-Vis: a visual analytic framework 
for Training and fine-Tuning Transformers
in NLP}

\author{Raymond Li\textsuperscript{*}, Wen Xiao\textsuperscript{†}, Lanjun Wang\textsuperscript{‡}\thanks{~~Corresponding author.}~, Hyeju Jang\textsuperscript{†}, Giuseppe Carenini\textsuperscript{†} \\
\textsuperscript{†}University of British Columbia, Vancouver, Canada \\
\{raymondl, xiaowen3, hyejuj, carenini\}@cs.ubc.ca
\\ \textsuperscript{‡}Huawei Cananda Technologies Co. Ltd., Burnaby, Canada \\
lanjun.wang@huawei.com
}

\begin{document}
\maketitle
\begin{abstract}
Transformers are the dominant architecture in NLP, but their training and fine-tuning is still very challenging.
In this paper, we present the design and implementation of a visual analytic framework for assisting researchers in such process, by providing them with valuable insights about the model's intrinsic properties and behaviours. Our framework offers an intuitive overview that allows the user to explore different facets of the model (e.g., hidden states, attention) through interactive visualization, and allows a suite of built-in algorithms that compute the importance of model components and different parts of the input sequence. 
Case studies and feedback from a user focus group indicate that the framework is useful, and suggest several improvements.

\end{abstract}

\begin{figure*}
    \centering
    \includegraphics[width=\linewidth]{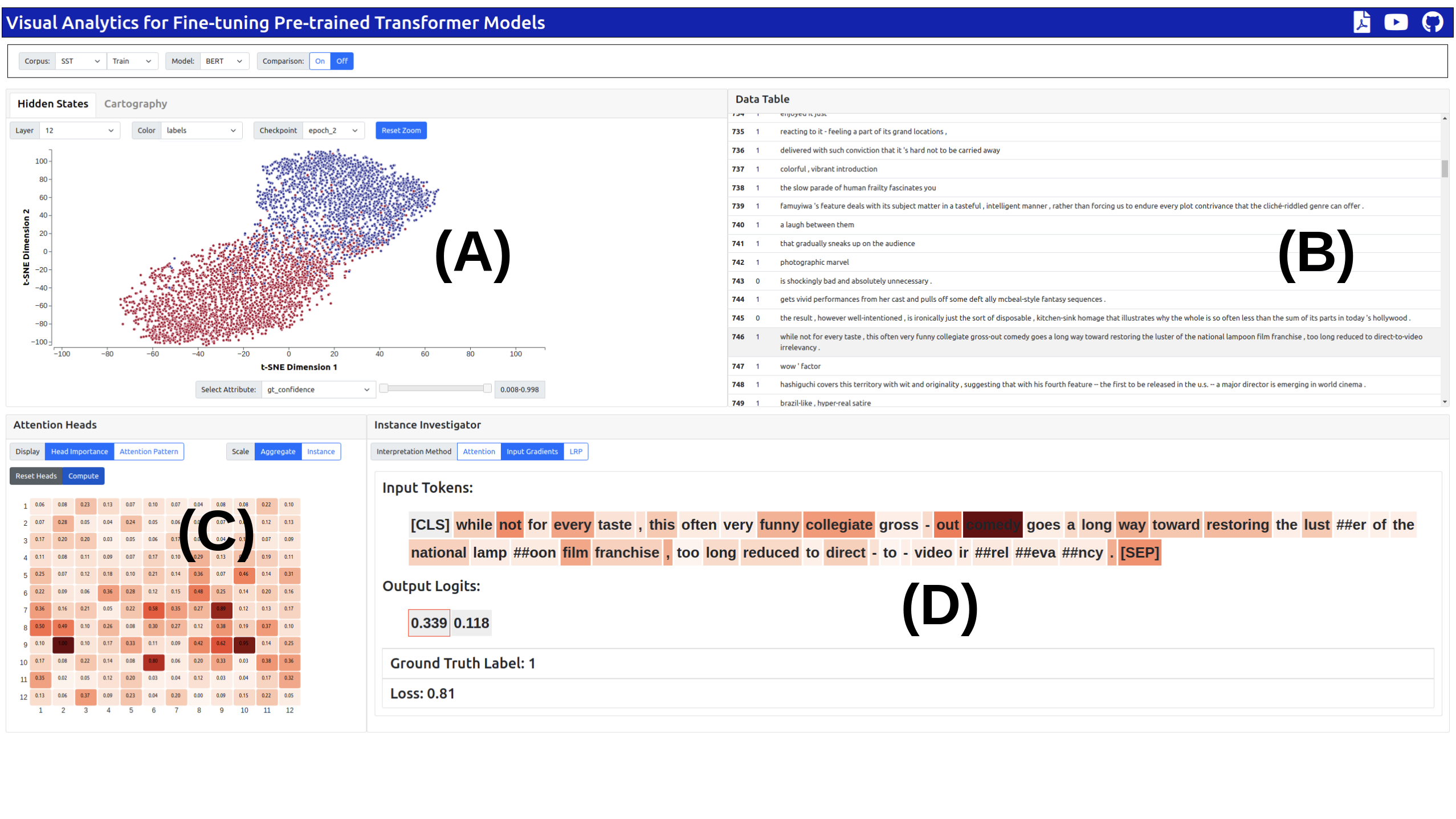}
    \vspace{-2.5em}
    \caption{Overview of the interface: \textbf{(A)} Projection View provides a 2D visualization of the dataset by encoding each example as a point on the scatterplot; \textbf{(B)} Data Table allows the user to view the content and metadata (e.g. label, loss) of the data examples (e.g. document); \textbf{(C)} Attention Head View visualizes the head importance and weight matrices of each attention head; \textbf{(D)} Instance Investigation View allows the user to perform detailed analysis (e.g. interpretation, attention) on a data example's input sequence.}
    \label{fig:teaser}
    \vspace{-2mm}
\end{figure*}


\section{Introduction}
Neural approaches 
have made significant progress in recent years, with Transformer models \citep{vaswani2017attention}  rapidly becoming the dominant architecture in NLP, due to their efficient parallel training and their ability to effectively capture features of long sequences. 
Following the release of BERT \citep{devlin-etal-2019-bert} along with other Transformer models pretrained on large corpora \citep{liu2019roberta, lewis-etal-2020-bart, joshi-etal-2020-spanbert, lee2020biobert}, the most successful strategy on many 
NLP leaderboards is currently to fine-tune such pretrained models to the particular target NLP  task (e.g., summarization, text classification).
However, despite the strong empirical performance of this 
strategy, understanding and interpreting the training and fine-tuning processes remains a critical and challenging step for model developers and researchers \citep{kovaleva-etal-2019-revealing, hao-etal-2019-visualizing, merchant-etal-2020-happens, hao-etal-2020-investigating}. 

Generally speaking, a large number of visual analytics tools have been shown to effectively support the analysis and interpretation of deep learning models 
\citep{hohman2018visual}. For instance, to remedy the black-box nature of neural network hidden states, previous work has used scatterplots to visualize high dimensional vectors in  projection views \citep{smilkov2016embedding, kahng2017cti}, with \citet{aken2020visbert} visualizing 
the differences of token representations from different layers of BERT \citep{devlin-etal-2019-bert}.
Similarly, despite some limitations
regarding the explanatory capabilities of the attention mechanism
\citep{jain-wallace-2019-attention, wiegreffe-pinter-2019-attention}, its visualization has also been shown to be beneficial, with promising recent work focusing on Transformers \citep{vig-2019-multiscale, hoover-etal-2020-exbert}.


Besides the works on exploring what has been learnt in the pretrained models, there are also several visualization tools developed
to show saliency scores generated by gradient-based \citep{simonyan2013deep, bach2015pixel, shrikumar2017learning} and perturbation-based methods \citep{ribeiro16, li2016understanding},
which can  help with interpreting the relative importance of individual tokens in the input with respect to a target prediction \citep{wallace-etal-2019-allennlp, johnson2020njm, tenney-etal-2020-language}.
However, 
only a few studies have instead focused on visualizing the overall training dynamics, where  support is critical for identifying mislabeled or failure cases \citep{liu2018deeptracker, xiang2019interactive, swayamdipta-etal-2020-dataset}

In essence, the T\textsuperscript{3}-Vis framework we propose in this paper synergistically integrates  some of the interactive visualizations mentioned above to support developers in the challenging task of training and fine-tuning Transformers. 
This is in contrast with other similar recent visual tools (\autoref{table:recent-comparison}), which either only focus on single data point explanations for uncovering model bias and finding decision boundaries (e.g., AllenNLP Interpret \citep{wallace-etal-2019-allennlp}), or only focus on analyzing failed examples and understanding model's behaviour (e.g., Language Interpretability Tool (LIT) \citep{tenney-etal-2020-language}).

\begin{table*}[ht!]
\begin{center}
\resizebox{\linewidth}{!}{
\begin{tabular}{ |c|c|c|c|c|c|c|c|c| } 
    \hline
    \multirow{2}{*}{\large{Frameworks}} & \multicolumn{5}{c|}{Components} & \multicolumn{3}{c|}{Functions} \\
    \cline{2-9}
    & Dataset & Embeddings & \begin{tabular}{@{}c@{}}Head \\ Importance\end{tabular} & Attention & \begin{tabular}{@{}c@{}}Training \\ Dynamics\end{tabular} & Interpretations & Pruning & Comparison \\
    \hline
    BertViz \cite{vig-2019-multiscale} & & & & \checkmark & & & & \\
    \hline
    \begin{tabular}{@{}c@{}}AllenNLP Interpret \\ \cite{wallace-etal-2019-allennlp}\end{tabular}  & & & & & & \checkmark & & \\
    \hline
    exBERT \cite{hoover-etal-2020-exbert} & \checkmark & \checkmark & & \checkmark & & & & \\
    \hline
    LIT \cite{tenney-etal-2020-language}  & \checkmark & \checkmark & & \checkmark & & \checkmark & & \checkmark \\
    \hline
    InterperT \citep{lal-etal-2021-interpret} & \checkmark & \checkmark & \checkmark & \checkmark & &  & & \\
    \hline
    \textbf{T\textsuperscript{3}-Vis} & \checkmark & \checkmark & \checkmark & \checkmark & \checkmark & \checkmark & \checkmark & \checkmark \\
    \hline
\end{tabular}
}
\end{center}
\caption{Comparison with other visual frameworks from recent work.}
\label{table:recent-comparison}
\end{table*}






Following 
the well-established Nested Model for visualization design \citep{munzner2009nested}, we first 
perform an extensive requirement analysis, from which 
we derive user tasks and data abstractions 
to guide the design of visual encoding and interaction techniques. More specifically, the resulting T\textsuperscript{3}-Vis framework provides an intuitive overview that allows users to explore different facets of the model (e.g., hidden states, attention, training dynamics) through interactive visualization, along with a suite of built-in algorithms that compute the importance of model components and different parts of the input sequence.

Our contributions are as follows: 
\textbf{(1)} An extensive user requirement analysis on supporting the training and fine-tuning of Transformer models, based on extensive literature review and interviews with NLP researchers,  
\textbf{(2)} the design and implementation of an open-sourced visual analytic framework for assisting researchers in the fine-tuning process 
with a suite of built-in interpretation methods for understanding model behaviour, and 
\textbf{(3)} the first steps of an iterative design based on case studies and feedback from a user focus group.

\section{Visualization Design}
The design of our T\textsuperscript{3}-Vis is based on the 
nested model for InfoVis design
\citep{munzner2009nested}. 

\subsection{User Requirements}
\label{ssec:user_requirements}
To derive useful analytical user tasks, 
we first identify a set of high-level user requirements through interviews with five NLP researchers as well as surveying recent literature related to the interpretability of pretrained Transformers. In the interviews, we prompt participants with the open-ended question of "\emph{If a visualization tool is provided to speed up your development (using fine-tuning pretrained Transformers), what information would you like to see and explore?}". 
Combining the interview results and insights from the literature review,
we organize these findings into five high-level requirements, each highlighting a different facet of the model for visualization.

\textbf{Hidden state visualization (UR-1)}: Support the exploration for hidden state representations from the pretrained model 
to assist users in the training process.

\textbf{Attention visualization (UR-2)}: 
Allow users 
to examine and explore the linguistic or positional patterns exhibited in the self-attention distribution for different attention heads in the model.

\textbf{Attention head importance(UR-3)}: 
Enable users to investigate and understand the importance of the attention heads for the downstream task and the effects of pruning them on the model's behaviour.

\textbf{Interpretability of models (UR-4)}:  In addition to attention maps, 
support a suite of alternative explanation methods based on token input importance, thus allowing users to better understand the model behaviours 
during inference.

\textbf{Training dynamics (UR-5)}:  
Assist users in identifying relevant data examples based on their roles in the training process. 

\begin{table*}[ht!]
\begin{center}
\resizebox{\linewidth}{!}{
\begin{tabular}
{ |c|c|c|c|c|c|c| } 
 \hline
 \textbf{\#} & \textbf{Question} & \textbf{When} & \textbf{Granularity} & \textbf{UR} & \textbf{Components} \\
 \hline
 T1 & \shortstack{How to determine the model \\ representation for a given NLP task?} & Before & Dataset & UR-1 & Projection \\
 \hline
 T2 & \shortstack{What are the outliers of the dataset?} & Before, During, After & Dataset & UR-1, UR-5 & Projection\\ 
 \hline
 T3 & \shortstack{What types of linguistic or positional attributes do \\ the attention patterns exhibit for each attention heads?}   \ & Before, During, After & Instance & UR-2 & Projection\\
 \hline
 T4 & \shortstack{Which attention heads are considered important \\ for the task, and what are its functions?
 } & After & Both & UR-2, UR-3 & \shortstack{Attention Head \\ Instance Investigator}\\ 
 \hline
 T5 & \shortstack{How does pruning attention heads affects the model?}  & After & Instance & UR-3 & Attention Head \\ 
 \hline
 T6 & \shortstack{How does the model changes at \\ different stages of fine-tuning?} & During, After & Both & UR-1, UR-2, UR-3 & All\\ 
 \hline
 T7 & \shortstack{Does the model rely on specific parts of the \\ input sequence when making predictions? } & After & Instance & UR-4 & Instance Investigator\\ 
 \hline
 T8 & \shortstack{Are there mislabeled examples in the dataset?} & During, After & Both & UR-1, UR-5 & Projection\\ 
 \hline
 T9 & \shortstack{How can the dataset be augmented to improve \\the performance and robustness of the model?}& During, After & Both & UR-5 & Projection \\ 
 \hline
\end{tabular}}
\end{center}
\caption{Supported analytical tasks: questions that our interface helps to answer.}
\label{table:task-abstractions}
\end{table*}

\subsection{Supported Tasks and Data Model}

Based on these user requirements, we derive nine analytical tasks framed as information seeking questions (\autoref{table:task-abstractions}). 
The questions are categorized based on their \textit{Granularity} (dataset vs. instance-level) and \textit{When} they are relevant during the fine-tuning process. 
If we then look at the specific data the tasks are applied to, we characterize our data model as comprising the model hidden states, the dataset examples along with their label/training features, the attention values, head importance scores, and input saliency map.
Although our task and data models are derived for the fine-tuning of pretrained models, they can naturally be extended to training any 
Transformer models from scratch.
Importantly, all the questions are invariant to any Transformer-based models 
for any downstream tasks (e.g. classification, sequence-generation or labeling).

\subsection{T\textsuperscript{3}-Vis Components: Visual Encoding and Interactive Techniques}
\label{ssec:components}
 
\begin{figure}
    \centering
    \includegraphics[width=\linewidth]{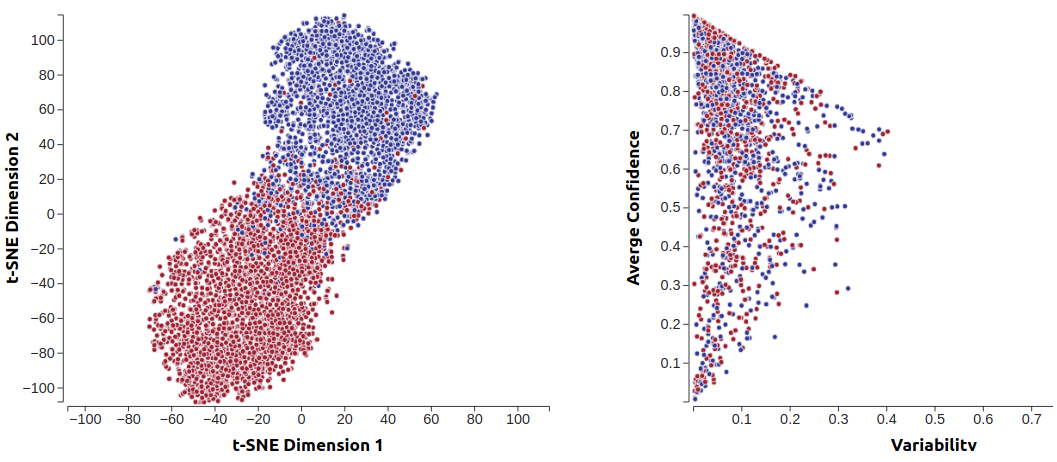}
    \caption{Interactive scatterplots based on the data examples' training dynamics (left), and the t-SNE projections of hidden states (right)}
    \label{fig:projection-view}
\end{figure}

\textbf{Projection View}
To assist users in visualizing the model's hidden state representation (\textbf{UR-1}) and to identify the training role of the data examples (\textbf{UR-5}), we design the Projection View (Figure \ref{fig:teaser}-(A)) as the main overview of our interface, and visualize the entire (or a subset of the) dataset on a 2D scatterplot, where each data point on the plot encodes a single data example (e.g. document) in the dataset. While the scatterplots can be generated in a variety of ways based on the user's needs, we provide two implementations (See Figure \ref{fig:projection-view}): (1) t-SNE projection \citep{van2008visualizing} of the model's hidden states, and (2) plotting the examples by their confidence and variability across epochs based on the Data Map technique \citep{swayamdipta-etal-2020-dataset}. The color of the data points can be selected by the user via a dropdown menu to encode attributes of the data examples, where color saturation is used for continuous attributes (e.g. loss, prediction confidence), while hue is used for categorical attributes (e.g. labels, prediction). The user can also filter the data points by
attributes, where a range slider is used for filtering the data points by continuous attributes, while a selectable dropdown menu is used to filter by categorical attributes. 
Furthermore, we also introduce a comparison mode by displaying the two scatterplots side-by-side, which allows for the flexibility of comparing across different checkpoints and the projection of different hidden state layers.

\textbf{Data Table}
The Data Table (Figure \ref{fig:teaser}-(B)) lists all examples of the dataset in a single scrollable list, where each row of the table shows the input text of a data example along with its ground truth label. When the user filters the dataset in the Projection View, the Data Table is also updated simultaneously, where only examples that satisfy the filtered criteria are kept.   

\begin{figure}[!tbp]
  \begin{subfigure}[b]{0.49\linewidth}
    \includegraphics[width=\linewidth]{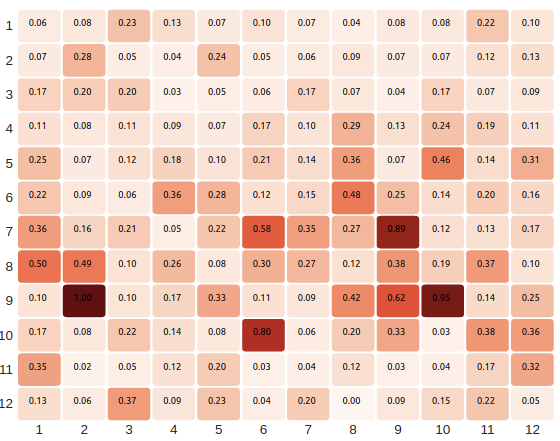}
    \caption{Head Importance}
    \label{fig:attention-importance}
  \end{subfigure}
  \hfill
  \begin{subfigure}[b]{0.49\linewidth}
    \includegraphics[width=\linewidth]{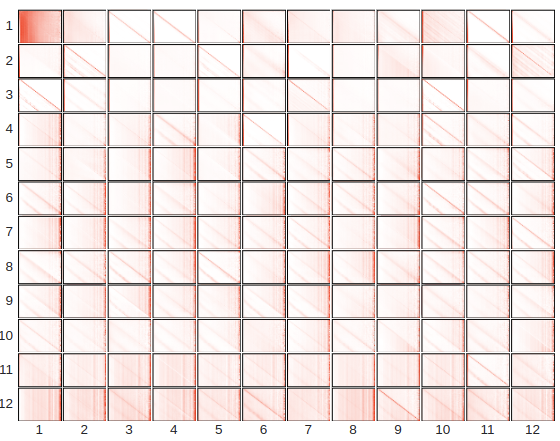}
    \caption{Attention Pattern}
    \label{fig:attention-weights}
  \end{subfigure}
  \caption{The two visualization techniques in the Attention Head View.}
  \label{fig:attention-head-view}
  \vspace{-2mm}
\end{figure}

\textbf{Attention Head View}
In order to visualize the importance of the models' attention heads (\textbf{UR-4}), as well as the patterns encoded in the attention weight matrices (\textbf{UR-3}), we design the Attention Head View (Figure \ref{fig:teaser}-(C)) with an $l \times h$ matrix ($l$ layers and $h$ heads), where each block in the matrix represents a single attention head at the respective index for layer and head. In this view, we provide two separate visualization techniques: namely (1) Head Importance (\autoref{fig:attention-importance}) and (2) Attention Pattern (\autoref{fig:attention-weights}), that can be switched using a toggle button. The Head Importance technique visualizes the normalized task-specific head importance score \footnote{Details are in \ref{sec:head-importance} of the Appendix}, where score of each head is encoded with the background color saturation of each block with the value also displayed in the middle. On the other hand, the Attention Pattern technique uses a heatmap to visualize the self-attention pattern of each head where the color saturation encodes magnitude of the associated weight matrices. We also provide a toggle button for the user to visualize the importance score and attention patterns on two scales, where the \textbf{aggregate}-scale visualizes the score and patterns averaged over the entire dataset, while the \textbf{instance}-scale visualizes the score and patterns for a selected data example. Lastly, we also offer an interactive technique for the user to dynamically prune attention heads and visualize the effects on a selected example. By hovering over each attention head block in the view, the user can click on the close icon to prune the respective attention head from the model.

\textbf{Instance Investigation View}
After the user selects a data example from the Projection View or Data Table, the Instance Investigation View (Figure \ref{fig:teaser}-(D)) renders the corresponding input text sequence along with the model predictions and labels to allow the user to perform detailed analysis on the data example. Our interface provides two analysis techniques: (1) self-attention weights (\textbf{UR-3}), and (2) input interpretation methods (\textbf{UR-4}). In this view, each token of the input sequence is displayed in a separate text block, where the background color saturation of each text block encodes the relative saliency or importance of the token based on the interpretation methods. 
By selecting a head in the Attention Head View (\autoref{fig:attention-head-view}), the user can click on the text block of any input token to visualize the self-attention distribution of the selected token over the input text sequence (what the selected token attends to). Similarly, the user can visualize the input saliency map
with respect to a model output, by clicking the corresponding output token. Our interface provides the implementation of two input interpretation methods\footnote{Details are in \ref{sec:input-interpretation} of the Appendix}
: (1) Layer-wise relevance propagation \citep{bach2015pixel}, and (2) input gradient \citep{simonyan2013deep}.

\subsection{Implementation}



\paragraph{Data Processing}
For each model checkpoint, data pertaining to dataset-level visualizations including hidden state projections, prediction confidence/variability, head importance score, and other attributes (e.g. loss, prediction) are first processed and saved in a back-end directory. The only added computational overhead to the user's training process is the dimension reduction algorithm for projecting hidden state representation, as other visualized values can all be extracted from the forward (e.g. confidence, variability, loss) and backward pass (e.g. head importance, input saliency) of model training.

\paragraph{Back-end}
Our back-end Python server provides built-in support for the PyTorch HuggingFace library \citep{wolf-etal-2020-transformers}, including methods for extracting attention values, head pruning, computing importance scores, and interpreting the model predictions. In order to avoid saving instance-level data (e.g., attention weights, input heatmap, etc.) for all examples in the dataset, our python server dynamically computes these values for a selected data example by performing a single forward and backward pass on the model. This requires the server to keep track of the model's current state, as well as a dataloader for indexing the selected data example. 

\paragraph{Front-end}
Our front-end implementation keeps track of the current visual state of the interface including the selections, filters, and checkpoint. The interface can be accessed through any web browser, where data is retrieved from the back-end server via RESTful API. The interactive visual components of the interface are implemented using the D3.js \citep{bostock2011d3}, and other UI components (e.g. buttons, sliders) are implemented with popular front-end libraries (e.g. jQuery, Bootstrap).  

\section{Iterative Design}
\subsection{Focus Group Study}
In order to collect suggestions and initial feedback on T\textsuperscript{3}-Vis, we conducted a focus group study with $20$ NLP researchers that work regularly with pretrained Transformer models. In this study, we first presented the design of the interface, then gave a demo showing its usage on an example, and throughout the process we gathered responses from the participants. 

Most positive feedback focused on the effectiveness of our techniques for visualizing self-attention especially on longer documents (in contrast to showing links between tokens \citep{vig-2019-multiscale}). There were also comments on the usefulness of the input saliency map in providing insightful clues on the model's decision process.

Some participants also suggested that the interface would be more useful for classification problems with well-defined evaluation metrics since data examples tended to be better clustered in the Projection View so that they could be easily filtered for error analysis.
The need of optimizing the front-end to support the visualization of large-scale datasets was also mentioned. 

On the negative side, some participants were concerned by the information loss intrinsic in the dimension reduction methods, whose possible negative effects on the user analysis tasks definitely requires further study. Encouragingly, at the end, a few participants expressed interest in applying and evaluating T\textsuperscript{3}-Vis on their datasets and NLP tasks.



\subsection{
Case Studies}
This section describes two case studies of how T\textsuperscript{3}-Vis facilitates the understanding and exploration of the fine-tuning process through applications with real-world corpora. These studies provide initial evidence on the effectiveness of different visualization components, and serve as examples for how our framework can be used in applications.

\subsubsection{Pattern Exploration for an Extractive Summarizer}

NLP researchers in our group, who work on summarization,  applied T\textsuperscript{3}-Vis to the extractive summarization task, which aims to compress a document by selecting its most informative sentences.
BERTSum, which is fine-tuned from a BERT model\cite{liu-lapata-2019-text}, is one of the top-performing models for extractive summarization,  but why and how it works remains a mystery. With our interface, the researchers explored patterns captured by the model that played important roles in model predictions. They performed an analysis on the CNN/Daily Mail dataset \cite{cnndm}, which is arguably the most popular benchmark for summarization tasks. 

The first step was to find the important heads among all the heads across all the layers. From the Head Importance View (Figure \ref{fig:teaser}-(C)), the researchers selected the attention heads with high head importance scores, so that the corresponding attention distribution was available to interact with. Then they selected some tokens in the Attention View to see which tokens they mostly attended to, and repeated this process for multiple other data examples, in order to explore whether there was a general pattern across different data examples. 

While examining attention heads based on their importance in descending order, the researchers observed that tokens tended to have high attention on other tokens of the same word on the important attention heads. For example, the token ``victim" attributed almost all of its attention score to other instances of the token ``victim" in the source document. They further found two more patterns in other important heads, in which the tokens tended to have more attention on the tokens within the same sentence, as well as the adjacent tokens.
These behaviours were consistent across different pretrained models (e.g. RoBERTaSum). 

These findings provided useful insights to assist the researchers in designing more efficient and accurate summarization models in the future, and served as a motivation for the researchers to perform similar analysis for other NLP tasks.

\subsubsection{Error Analysis for Topic Classification}
Other researchers in our group explored the interface for error analysis to identify possible improvements of a BERT-based model for topic classification. The Yahoo Answers dataset \citep{zhang2015character} was used, which contains 10 topic classes. 

Researchers first used the Projection View (Figure \ref{fig:teaser}-(A)) to find misclassified data examples as applying filters to select label and prediction classes. For a selected topic class in the t-SNE projection of the model's hidden states, they found out that the misclassified data points far away from clusters of correctly predicted examples were often mislabeled during annotation. Therefore, misclassfied data points within such clusters were of greater interest to them since such points tends to indicate model failures (instead of mistakes in annotation). 
Furthermore, data points in the area with low variability and low confidence on the Data Map plot were also selected for investigation since they are interpreted as consistently misclassified across epochs. After selecting the examples, the researchers inspected each instance by using the Instance Investigation View (Figure \ref{fig:teaser}-(D)) with the Input Gradient method to visualize the input saliency map for the prediction of each class. 


From this analysis, they discovered two scenarios that led to misclassification. First, the model focused on unimportant and possibly misleading details that are not representative of the document's overall topic. For instance, a document about \textit{Business \& Finance} was classified into the \textit{Sport} category because the model attended to ``hockey player'', ``football player'', and ``baseball player'', which were listed as job titles while discussing available jobs in Michigan. Second, the model failed in cases where background knowledge is required. For example, a document under the \textit{Entertainment \& Music} category mentioned names of two actors which were key clues for the topic, but the model only attended to other words, and made a wrong prediction.

These findings helped researchers to gain insights for future model design where additional information such as discourse structure (which can better reveal importance) and encyclopedic knowledge could be injected into the model's architecture to improve the task performance.

\section{Conclusion}
In this paper, we presented T\textsuperscript{3}-Vis, a visual analytic framework designed to help researchers better understand training and fine-tuning processes of  Transformer-based models. Our visual interface provides faceted visualization of a Transformer model and allows exploring data across multiple granularities, while enabling users to dynamically interact with the model. 
Our focus group and case studies demonstrated the effectiveness of our interface by assisting the researchers in interpreting the models' behaviour and identifying potential directions to improve task performances. 

For future work, we will continue to improve our framework through the iterative process of exploring further usage scenarios and collecting feedback from users. We will extend our framework to provide a more advanced visualization for custom Transformers. For example, we may want to support the visualization of models with more complex connections (e.g. parallel attention layers) or an advanced attention mechanism (e.g. sparse attention). 



\clearpage

\bibliography{anthology,custom}

\begin{thebibliography}{40}
\expandafter\ifx\csname natexlab\endcsname\relax\def\natexlab#1{#1}\fi

\bibitem[{Aken et~al.(2020)Aken, Winter, L{\"o}ser, and Gers}]{aken2020visbert}
Betty~van Aken, Benjamin Winter, Alexander L{\"o}ser, and Felix~A Gers. 2020.
\newblock Visbert: Hidden-state visualizations for transformers.
\newblock In \emph{Companion Proceedings of the Web Conference 2020}, pages
  207--211.

\bibitem[{Bach et~al.(2015)Bach, Binder, Montavon, Klauschen, M{\"u}ller, and
  Samek}]{bach2015pixel}
Sebastian Bach, Alexander Binder, Gr{\'e}goire Montavon, Frederick Klauschen,
  Klaus-Robert M{\"u}ller, and Wojciech Samek. 2015.
\newblock On pixel-wise explanations for non-linear classifier decisions by
  layer-wise relevance propagation.
\newblock \emph{PloS one}, 10(7).

\bibitem[{Bostock et~al.(2011)Bostock, Ogievetsky, and Heer}]{bostock2011d3}
Michael Bostock, Vadim Ogievetsky, and Jeffrey Heer. 2011.
\newblock D$^3$ data-driven documents.
\newblock \emph{IEEE transactions on visualization and computer graphics},
  17(12):2301--2309.

\bibitem[{Devlin et~al.(2019)Devlin, Chang, Lee, and
  Toutanova}]{devlin-etal-2019-bert}
Jacob Devlin, Ming-Wei Chang, Kenton Lee, and Kristina Toutanova. 2019.
\newblock \href {https://doi.org/10.18653/v1/N19-1423} {{BERT}: Pre-training of
  deep bidirectional transformers for language understanding}.
\newblock In \emph{Proceedings of the 2019 Conference of the North {A}merican
  Chapter of the Association for Computational Linguistics: Human Language
  Technologies, Volume 1 (Long and Short Papers)}, pages 4171--4186,
  Minneapolis, Minnesota. Association for Computational Linguistics.

\bibitem[{Hao et~al.(2019)Hao, Dong, Wei, and Xu}]{hao-etal-2019-visualizing}
Yaru Hao, Li~Dong, Furu Wei, and Ke~Xu. 2019.
\newblock \href {https://doi.org/10.18653/v1/D19-1424} {Visualizing and
  understanding the effectiveness of {BERT}}.
\newblock In \emph{Proceedings of the 2019 Conference on Empirical Methods in
  Natural Language Processing and the 9th International Joint Conference on
  Natural Language Processing (EMNLP-IJCNLP)}, pages 4143--4152, Hong Kong,
  China. Association for Computational Linguistics.

\bibitem[{Hao et~al.(2020)Hao, Dong, Wei, and Xu}]{hao-etal-2020-investigating}
Yaru Hao, Li~Dong, Furu Wei, and Ke~Xu. 2020.
\newblock \href {https://www.aclweb.org/anthology/2020.aacl-main.11}
  {Investigating learning dynamics of {BERT} fine-tuning}.
\newblock In \emph{Proceedings of the 1st Conference of the Asia-Pacific
  Chapter of the Association for Computational Linguistics and the 10th
  International Joint Conference on Natural Language Processing}, pages 87--92,
  Suzhou, China. Association for Computational Linguistics.

\bibitem[{Hermann et~al.(2015)Hermann, Kocisky, Grefenstette, Espeholt, Kay,
  Suleyman, and Blunsom}]{cnndm}
Karl~Moritz Hermann, Tomas Kocisky, Edward Grefenstette, Lasse Espeholt, Will
  Kay, Mustafa Suleyman, and Phil Blunsom. 2015.
\newblock \href
  {https://proceedings.neurips.cc/paper/2015/file/afdec7005cc9f14302cd0474fd0f3c96-Paper.pdf}
  {Teaching machines to read and comprehend}.
\newblock In \emph{Advances in Neural Information Processing Systems},
  volume~28, pages 1693--1701. Curran Associates, Inc.

\bibitem[{Hohman et~al.(2018)Hohman, Kahng, Pienta, and
  Chau}]{hohman2018visual}
Fred Hohman, Minsuk Kahng, Robert Pienta, and Duen~Horng Chau. 2018.
\newblock Visual analytics in deep learning: An interrogative survey for the
  next frontiers.
\newblock \emph{IEEE transactions on visualization and computer graphics},
  25(8):2674--2693.

\bibitem[{Hoover et~al.(2020)Hoover, Strobelt, and
  Gehrmann}]{hoover-etal-2020-exbert}
Benjamin Hoover, Hendrik Strobelt, and Sebastian Gehrmann. 2020.
\newblock \href {https://doi.org/10.18653/v1/2020.acl-demos.22} {ex{BERT}: {A}
  {V}isual {A}nalysis {T}ool to {E}xplore {L}earned {R}epresentations in
  {T}ransformer {M}odels}.
\newblock In \emph{Proceedings of the 58th Annual Meeting of the Association
  for Computational Linguistics: System Demonstrations}, pages 187--196,
  Online. Association for Computational Linguistics.

\bibitem[{Jain and Wallace(2019)}]{jain-wallace-2019-attention}
Sarthak Jain and Byron~C. Wallace. 2019.
\newblock \href {https://doi.org/10.18653/v1/N19-1357} {{A}ttention is not
  {E}xplanation}.
\newblock In \emph{Proceedings of the 2019 Conference of the North {A}merican
  Chapter of the Association for Computational Linguistics: Human Language
  Technologies, Volume 1 (Long and Short Papers)}, pages 3543--3556,
  Minneapolis, Minnesota. Association for Computational Linguistics.

\bibitem[{Johnson et~al.(2020)Johnson, Carenini, and Murray}]{johnson2020njm}
David Johnson, Giuseppe Carenini, and Gabriel Murray. 2020.
\newblock Njm-vis: interpreting neural joint models in nlp.
\newblock In \emph{Proceedings of the 25th International Conference on
  Intelligent User Interfaces}, pages 286--296.

\bibitem[{Joshi et~al.(2020)Joshi, Chen, Liu, Weld, Zettlemoyer, and
  Levy}]{joshi-etal-2020-spanbert}
Mandar Joshi, Danqi Chen, Yinhan Liu, Daniel~S. Weld, Luke Zettlemoyer, and
  Omer Levy. 2020.
\newblock \href {https://doi.org/10.1162/tacl_a_00300} {{S}pan{BERT}: Improving
  pre-training by representing and predicting spans}.
\newblock \emph{Transactions of the Association for Computational Linguistics},
  8:64--77.

\bibitem[{Kahng et~al.(2017)Kahng, Andrews, Kalro, and Chau}]{kahng2017cti}
Minsuk Kahng, Pierre~Y Andrews, Aditya Kalro, and Duen Horng~Polo Chau. 2017.
\newblock Activis: Visual exploration of industry-scale deep neural network
  models.
\newblock \emph{IEEE transactions on visualization and computer graphics},
  24(1):88--97.

\bibitem[{Kovaleva et~al.(2019)Kovaleva, Romanov, Rogers, and
  Rumshisky}]{kovaleva-etal-2019-revealing}
Olga Kovaleva, Alexey Romanov, Anna Rogers, and Anna Rumshisky. 2019.
\newblock \href {https://doi.org/10.18653/v1/D19-1445} {Revealing the dark
  secrets of {BERT}}.
\newblock In \emph{Proceedings of the 2019 Conference on Empirical Methods in
  Natural Language Processing and the 9th International Joint Conference on
  Natural Language Processing (EMNLP-IJCNLP)}, pages 4365--4374, Hong Kong,
  China. Association for Computational Linguistics.

\bibitem[{Lal et~al.(2021)Lal, Ma, Aflalo, Howard, Simoes, Korat, Pereg,
  Singer, and Wasserblat}]{lal-etal-2021-interpret}
Vasudev Lal, Arden Ma, Estelle Aflalo, Phillip Howard, Ana Simoes, Daniel
  Korat, Oren Pereg, Gadi Singer, and Moshe Wasserblat. 2021.
\newblock \href {https://www.aclweb.org/anthology/2021.eacl-demos.17}
  {{I}nterpre{T}: An interactive visualization tool for interpreting
  transformers}.
\newblock In \emph{Proceedings of the 16th Conference of the European Chapter
  of the Association for Computational Linguistics: System Demonstrations},
  pages 135--142, Online. Association for Computational Linguistics.

\bibitem[{Lee et~al.(2020)Lee, Yoon, Kim, Kim, Kim, So, and
  Kang}]{lee2020biobert}
Jinhyuk Lee, Wonjin Yoon, Sungdong Kim, Donghyeon Kim, Sunkyu Kim, Chan~Ho So,
  and Jaewoo Kang. 2020.
\newblock Biobert: a pre-trained biomedical language representation model for
  biomedical text mining.
\newblock \emph{Bioinformatics}, 36(4):1234--1240.

\bibitem[{Lewis et~al.(2020)Lewis, Liu, Goyal, Ghazvininejad, Mohamed, Levy,
  Stoyanov, and Zettlemoyer}]{lewis-etal-2020-bart}
Mike Lewis, Yinhan Liu, Naman Goyal, Marjan Ghazvininejad, Abdelrahman Mohamed,
  Omer Levy, Veselin Stoyanov, and Luke Zettlemoyer. 2020.
\newblock \href {https://doi.org/10.18653/v1/2020.acl-main.703} {{BART}:
  Denoising sequence-to-sequence pre-training for natural language generation,
  translation, and comprehension}.
\newblock In \emph{Proceedings of the 58th Annual Meeting of the Association
  for Computational Linguistics}, pages 7871--7880, Online. Association for
  Computational Linguistics.

\bibitem[{Li et~al.(2016)Li, Monroe, and Jurafsky}]{li2016understanding}
Jiwei Li, Will Monroe, and Dan Jurafsky. 2016.
\newblock Understanding neural networks through representation erasure.
\newblock \emph{arXiv preprint arXiv:1612.08220}.

\bibitem[{Liu et~al.(2018)Liu, Cui, Jin, Guo, and Qu}]{liu2018deeptracker}
Dongyu Liu, Weiwei Cui, Kai Jin, Yuxiao Guo, and Huamin Qu. 2018.
\newblock Deeptracker: Visualizing the training process of convolutional neural
  networks.
\newblock \emph{ACM Transactions on Intelligent Systems and Technology (TIST)},
  10(1):1--25.

\bibitem[{Liu and Lapata(2019)}]{liu-lapata-2019-text}
Yang Liu and Mirella Lapata. 2019.
\newblock \href {https://doi.org/10.18653/v1/D19-1387} {Text summarization with
  pretrained encoders}.
\newblock In \emph{Proceedings of the 2019 Conference on Empirical Methods in
  Natural Language Processing and the 9th International Joint Conference on
  Natural Language Processing (EMNLP-IJCNLP)}, pages 3730--3740, Hong Kong,
  China. Association for Computational Linguistics.

\bibitem[{Liu et~al.(2019)Liu, Ott, Goyal, Du, Joshi, Chen, Levy, Lewis,
  Zettlemoyer, and Stoyanov}]{liu2019roberta}
Yinhan Liu, Myle Ott, Naman Goyal, Jingfei Du, Mandar Joshi, Danqi Chen, Omer
  Levy, Mike Lewis, Luke Zettlemoyer, and Veselin Stoyanov. 2019.
\newblock Roberta: A robustly optimized bert pretraining approach.
\newblock \emph{arXiv preprint arXiv:1907.11692}.

\bibitem[{Merchant et~al.(2020)Merchant, Rahimtoroghi, Pavlick, and
  Tenney}]{merchant-etal-2020-happens}
Amil Merchant, Elahe Rahimtoroghi, Ellie Pavlick, and Ian Tenney. 2020.
\newblock \href {https://doi.org/10.18653/v1/2020.blackboxnlp-1.4} {What
  happens to {BERT} embeddings during fine-tuning?}
\newblock In \emph{Proceedings of the Third BlackboxNLP Workshop on Analyzing
  and Interpreting Neural Networks for NLP}, pages 33--44, Online. Association
  for Computational Linguistics.

\bibitem[{Molchanov et~al.(2019)Molchanov, Mallya, Tyree, Frosio, and
  Kautz}]{molchanov2019importance}
Pavlo Molchanov, Arun Mallya, Stephen Tyree, Iuri Frosio, and Jan Kautz. 2019.
\newblock Importance estimation for neural network pruning.
\newblock In \emph{Proceedings of the IEEE/CVF Conference on Computer Vision
  and Pattern Recognition}, pages 11264--11272.

\bibitem[{Montavon et~al.(2018)Montavon, Samek, and
  M{\"u}ller}]{montavon2018methods}
Gr{\'e}goire Montavon, Wojciech Samek, and Klaus-Robert M{\"u}ller. 2018.
\newblock Methods for interpreting and understanding deep neural networks.
\newblock \emph{Digital Signal Processing}, 73:1--15.

\bibitem[{Munzner(2009)}]{munzner2009nested}
Tamara Munzner. 2009.
\newblock A nested model for visualization design and validation.
\newblock \emph{IEEE transactions on visualization and computer graphics},
  15(6):921--928.

\bibitem[{Ribeiro et~al.(2016)Ribeiro, Singh, and Guestrin}]{ribeiro16}
Marco~Tulio Ribeiro, Sameer Singh, and Carlos Guestrin. 2016.
\newblock " why should i trust you?" explaining the predictions of any
  classifier.
\newblock In \emph{Proceedings of the 22nd ACM SIGKDD international conference
  on knowledge discovery and data mining}, pages 1135--1144.

\bibitem[{Shrikumar et~al.(2017)Shrikumar, Greenside, and
  Kundaje}]{shrikumar2017learning}
Avanti Shrikumar, Peyton Greenside, and Anshul Kundaje. 2017.
\newblock Learning important features through propagating activation
  differences.
\newblock In \emph{International Conference on Machine Learning}, pages
  3145--3153. PMLR.

\bibitem[{Simonyan et~al.(2013)Simonyan, Vedaldi, and
  Zisserman}]{simonyan2013deep}
Karen Simonyan, Andrea Vedaldi, and Andrew Zisserman. 2013.
\newblock Deep inside convolutional networks: Visualising image classification
  models and saliency maps.
\newblock \emph{arXiv preprint arXiv:1312.6034}.

\bibitem[{Smilkov et~al.(2016)Smilkov, Thorat, Nicholson, Reif, Vi{\'e}gas, and
  Wattenberg}]{smilkov2016embedding}
Daniel Smilkov, Nikhil Thorat, Charles Nicholson, Emily Reif, Fernanda~B
  Vi{\'e}gas, and Martin Wattenberg. 2016.
\newblock Embedding projector: Interactive visualization and interpretation of
  embeddings.
\newblock \emph{arXiv preprint arXiv:1611.05469}.

\bibitem[{Swayamdipta et~al.(2020)Swayamdipta, Schwartz, Lourie, Wang,
  Hajishirzi, Smith, and Choi}]{swayamdipta-etal-2020-dataset}
Swabha Swayamdipta, Roy Schwartz, Nicholas Lourie, Yizhong Wang, Hannaneh
  Hajishirzi, Noah~A. Smith, and Yejin Choi. 2020.
\newblock \href {https://doi.org/10.18653/v1/2020.emnlp-main.746} {Dataset
  cartography: Mapping and diagnosing datasets with training dynamics}.
\newblock In \emph{Proceedings of the 2020 Conference on Empirical Methods in
  Natural Language Processing (EMNLP)}, pages 9275--9293, Online. Association
  for Computational Linguistics.

\bibitem[{Tenney et~al.(2020)Tenney, Wexler, Bastings, Bolukbasi, Coenen,
  Gehrmann, Jiang, Pushkarna, Radebaugh, Reif, and
  Yuan}]{tenney-etal-2020-language}
Ian Tenney, James Wexler, Jasmijn Bastings, Tolga Bolukbasi, Andy Coenen,
  Sebastian Gehrmann, Ellen Jiang, Mahima Pushkarna, Carey Radebaugh, Emily
  Reif, and Ann Yuan. 2020.
\newblock \href {https://doi.org/10.18653/v1/2020.emnlp-demos.15} {The language
  interpretability tool: Extensible, interactive visualizations and analysis
  for {NLP} models}.
\newblock In \emph{Proceedings of the 2020 Conference on Empirical Methods in
  Natural Language Processing: System Demonstrations}, pages 107--118, Online.
  Association for Computational Linguistics.

\bibitem[{Van~der Maaten and Hinton(2008)}]{van2008visualizing}
Laurens Van~der Maaten and Geoffrey Hinton. 2008.
\newblock Visualizing data using t-sne.
\newblock \emph{Journal of machine learning research}, 9(11).

\bibitem[{Vaswani et~al.(2017)Vaswani, Shazeer, Parmar, Uszkoreit, Jones,
  Gomez, Kaiser, and Polosukhin}]{vaswani2017attention}
Ashish Vaswani, Noam Shazeer, Niki Parmar, Jakob Uszkoreit, Llion Jones,
  Aidan~N Gomez, {\L}ukasz Kaiser, and Illia Polosukhin. 2017.
\newblock Attention is all you need.
\newblock In \emph{Advances in neural information processing systems}, pages
  5998--6008.

\bibitem[{Vig(2019)}]{vig-2019-multiscale}
Jesse Vig. 2019.
\newblock \href {https://doi.org/10.18653/v1/P19-3007} {A multiscale
  visualization of attention in the transformer model}.
\newblock In \emph{Proceedings of the 57th Annual Meeting of the Association
  for Computational Linguistics: System Demonstrations}, pages 37--42,
  Florence, Italy. Association for Computational Linguistics.

\bibitem[{Voita et~al.(2019)Voita, Talbot, Moiseev, Sennrich, and
  Titov}]{voita-etal-2019-analyzing}
Elena Voita, David Talbot, Fedor Moiseev, Rico Sennrich, and Ivan Titov. 2019.
\newblock \href {https://doi.org/10.18653/v1/P19-1580} {Analyzing multi-head
  self-attention: Specialized heads do the heavy lifting, the rest can be
  pruned}.
\newblock In \emph{Proceedings of the 57th Annual Meeting of the Association
  for Computational Linguistics}, pages 5797--5808, Florence, Italy.
  Association for Computational Linguistics.

\bibitem[{Wallace et~al.(2019)Wallace, Tuyls, Wang, Subramanian, Gardner, and
  Singh}]{wallace-etal-2019-allennlp}
Eric Wallace, Jens Tuyls, Junlin Wang, Sanjay Subramanian, Matt Gardner, and
  Sameer Singh. 2019.
\newblock \href {https://doi.org/10.18653/v1/D19-3002} {{A}llen{NLP} interpret:
  A framework for explaining predictions of {NLP} models}.
\newblock In \emph{Proceedings of the 2019 Conference on Empirical Methods in
  Natural Language Processing and the 9th International Joint Conference on
  Natural Language Processing (EMNLP-IJCNLP): System Demonstrations}, pages
  7--12, Hong Kong, China. Association for Computational Linguistics.

\bibitem[{Wiegreffe and Pinter(2019)}]{wiegreffe-pinter-2019-attention}
Sarah Wiegreffe and Yuval Pinter. 2019.
\newblock \href {https://doi.org/10.18653/v1/D19-1002} {Attention is not not
  explanation}.
\newblock In \emph{Proceedings of the 2019 Conference on Empirical Methods in
  Natural Language Processing and the 9th International Joint Conference on
  Natural Language Processing (EMNLP-IJCNLP)}, pages 11--20, Hong Kong, China.
  Association for Computational Linguistics.

\bibitem[{Wolf et~al.(2020)Wolf, Debut, Sanh, Chaumond, Delangue, Moi, Cistac,
  Rault, Louf, Funtowicz, Davison, Shleifer, von Platen, Ma, Jernite, Plu, Xu,
  Le~Scao, Gugger, Drame, Lhoest, and Rush}]{wolf-etal-2020-transformers}
Thomas Wolf, Lysandre Debut, Victor Sanh, Julien Chaumond, Clement Delangue,
  Anthony Moi, Pierric Cistac, Tim Rault, Remi Louf, Morgan Funtowicz, Joe
  Davison, Sam Shleifer, Patrick von Platen, Clara Ma, Yacine Jernite, Julien
  Plu, Canwen Xu, Teven Le~Scao, Sylvain Gugger, Mariama Drame, Quentin Lhoest,
  and Alexander Rush. 2020.
\newblock \href {https://doi.org/10.18653/v1/2020.emnlp-demos.6} {Transformers:
  State-of-the-art natural language processing}.
\newblock In \emph{Proceedings of the 2020 Conference on Empirical Methods in
  Natural Language Processing: System Demonstrations}, pages 38--45, Online.
  Association for Computational Linguistics.

\bibitem[{Xiang et~al.(2019)Xiang, Ye, Xia, Wu, Chen, and
  Liu}]{xiang2019interactive}
Shouxing Xiang, Xi~Ye, Jiazhi Xia, Jing Wu, Yang Chen, and Shixia Liu. 2019.
\newblock Interactive correction of mislabeled training data.
\newblock In \emph{2019 IEEE Conference on Visual Analytics Science and
  Technology (VAST)}, pages 57--68. IEEE.

\bibitem[{Zhang et~al.(2015)Zhang, Zhao, and LeCun}]{zhang2015character}
Xiang Zhang, Junbo Zhao, and Yann LeCun. 2015.
\newblock Character-level convolutional networks for text classification.
\newblock \emph{Advances in neural information processing systems},
  28:649--657.

\end{thebibliography}
\bibliographystyle{acl_natbib}

\clearpage

\appendix
\section{Appendix}
\label{sec:appendix}

\subsection{Head Importance Score}
\label{sec:head-importance}
Although the multi-head self attention mechanism in Transformers allows the model to learn multiple types of relationships between input representations across a single hidden layer, the importance of the individual attention heads can vary depending on the downstream tasks. Following previous work, we adapt the Taylor expansion method \citep{molchanov2019importance} to estimate the error induced from removing a group of parameters from the model. In our implementation, we use the first-order expansion to avoid the overhead from computing the Hessian, where the gradient with respect to validation loss is summed over all parameters of an attention head to estimate its importance.

\subsection{Input Interpretation}
\label{sec:input-interpretation}
\paragraph{Input Gradients}
The input gradient method \citep{simonyan2013deep} computes the gradient with respect to each token. During inference, the class-score derivative can be computed through back-propagation. The saliency of the token $x_i$ for class $c$ of output $y$ could therefore be estimated using the first-order Taylor expansion $\frac{\partial y_c}{\partial x_i}x_i$.

\paragraph{Layer-wise Relevance Propagation}
Layer-wise Relevance Propagation (LRP) \citep{bach2015pixel} was originally proposed to visualize the contributions of single pixels to predictions for an image classifier. By recursively computing relevance from the output layer to the input layer, LRP is demonstrated to be useful in unravelling the inference process of neural networks and has been adopted in recent work to analyze Transformer models \citep{voita-etal-2019-analyzing}. The intuition behind LRP is that, each neuron of the network is contributed by neurons in the previous layer, and the total amount of contributions for each layer should be a constant during back-propagating, which is called the \emph{conservation principle}. LRP offers flexibility to design propagation rules to explain various deep neural networks, one example propagation rule is shown as follows \citep{montavon2018methods},
\begin{equation}
    R_i = \Sigma_j \frac{a_i w_{ij}}{\Sigma_{i}a_iw_{ij}}R_j
\end{equation}
where $R_i$ and $R_j$ are relevance scores of two neurons in consecutive layers, $a_i$ is the respective activation for neuron $i$, and $w_{ij}$ is the weight between neuron $i$ and $j$.

\end{document}